\documentclass[conference,compsoc]{IEEEtran}
\usepackage{graphicx}
\usepackage{amssymb}
\usepackage{amsmath}
\usepackage{booktabs}

\usepackage{lineno}
\usepackage{algorithm}
\usepackage{algorithmic}
\usepackage{multirow}

\ifCLASSINFOpdf
\else
\fi
\ifCLASSOPTIONcompsoc
 \usepackage[caption=false,font=footnotesize,labelfont=sf,textfont=sf]{subfig}
\else
 \usepackage[caption=false,font=footnotesize]{subfig}
\fi
\begin{document}
%
\title{Cooperative Cross-Stream Network for Discriminative Action Representation}

\author{\IEEEauthorblockN{Jingran~Zhang}
\IEEEauthorblockA{Center for Future Multimedia\\
School of Computer Science and Engineering\\
University of Electronic Science and Technology of China\\
Chengdu 610051, China\\
jrzhang339@gmail.com}
\and
\IEEEauthorblockN{Fumin~Shen}
\IEEEauthorblockA{Center for Future Multimedia\\
School of Computer Science and Engineering\\
University of Electronic Science and Technology of China\\
Chengdu 610051, China\\
fumin.shen@gmail.com}
\and
\IEEEauthorblockN{Xing~Xu}
\IEEEauthorblockA{Center for Future Multimedia\\
School of Computer Science and Engineering\\
University of Electronic Science and Technology of China\\
Chengdu 610051, China\\
xing.xu@uestc.edu.cn}
\and
\IEEEauthorblockN{Heng~Tao~Shen}
\IEEEauthorblockA{Center for Future Multimedia\\
School of Computer Science and Engineering\\
University of Electronic Science and Technology of China\\
Chengdu 610051, China\\
shenhengtao@hotmail.com}
}


%


\maketitle

\begin{abstract}
Spatial and temporal stream model has gained great success in video action recognition. Most existing works pay more attention to designing effective features fusion methods, which train the two-stream model in a separate way. However, it's hard to ensure discriminability and explore complementary information between different streams in existing works. In this work, we propose a novel cooperative cross-stream network that investigates the conjoint information in multiple different modalities. The jointly spatial and temporal stream networks feature extraction is accomplished by an end-to-end learning manner. It extracts this complementary information of different modality from a connection block, which aims at exploring correlations of different stream features. Furthermore, different from the conventional ConvNet that learns the deep separable features with only one cross entropy loss, our proposed model enhances the discriminative power of the deeply learned features and reduces the undesired modality discrepancy by jointly optimizing a modality ranking constraint and a cross entropy loss for both homogeneous and heterogeneous modalities. The modality ranking constraint constitute intra-modality discriminative embedding and inter-modality triplet constraint, and it reduces both the intra-modality and cross-modality feature variations. Experiments on three benchmark datasets demonstrate that by cooperating appearance and motion feature extraction, our method can achieve state-of-the-art or competitive performance compared with existing results.
\end{abstract}


%
\IEEEpeerreviewmaketitle

\section{Introduction}
\label{S:1}

\begin{figure}[t]
    \centering
    \includegraphics[width=0.8\columnwidth]{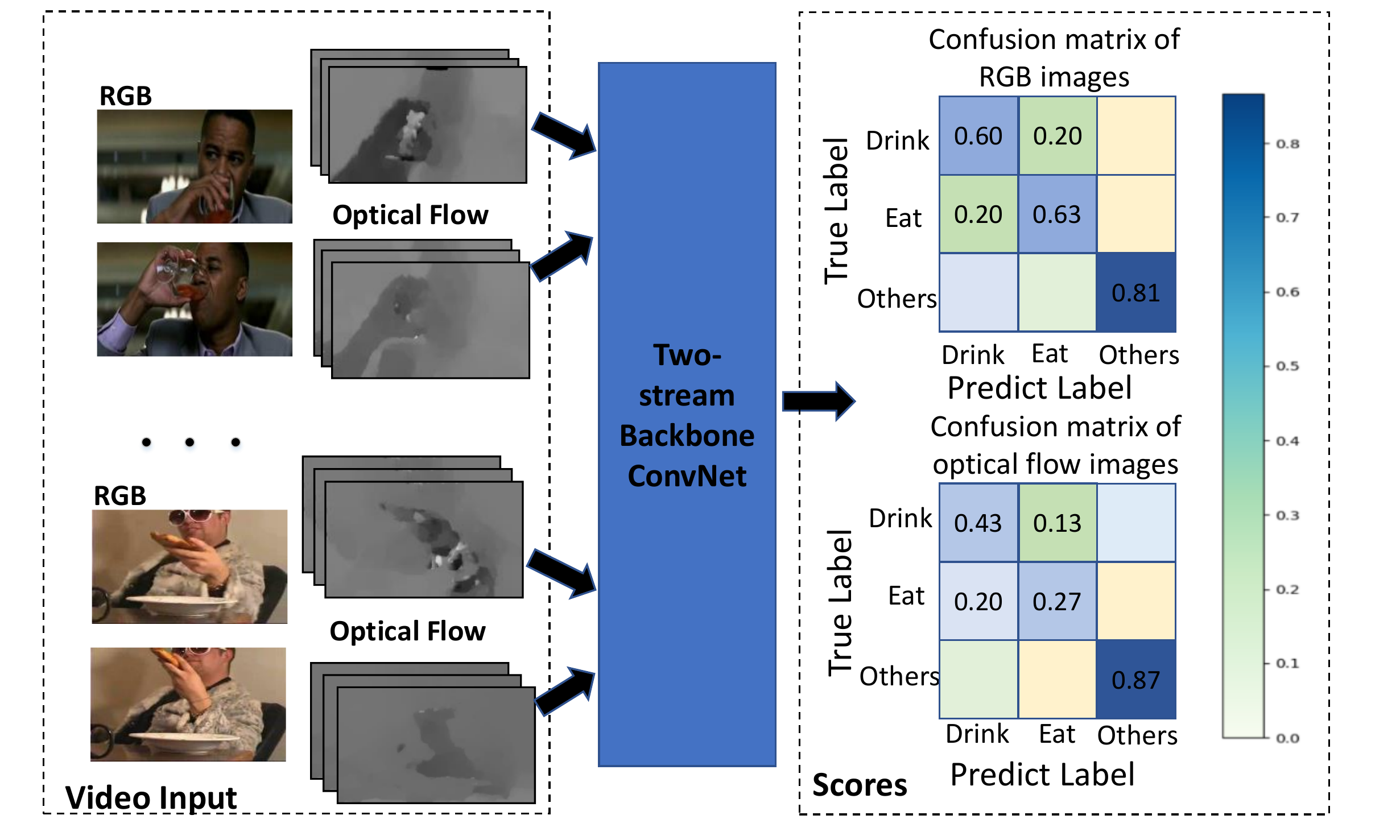}
    \caption{Traditional two-stream network for classifying ``drink'' and ``eat'' from HMDB-51 \cite{kuehne2013hmdb51} dataset. The input consist of video frames and optical flow images. It is hard to tell the classes when input the frames or optical flow field to TSN \cite{wang2016temporal} separately. The scores part is result of confusion matrix, which is derived from experiment of TSN on HMDB-51 dataset.}
    \label{fig:example}
\end{figure}

Video analysis has attracted significant attention from the academic community in computer vision, partly due to the rapidly growing number of videos being shared on the Internet. As one of the fundamental task in video analysis, human action recognition in videos is a well-studied problem. However, traditionary CNN-based representations \cite{karpathy2014large} have not yet significantly made as the transformation of an impact on action representation as it does on still images, because of significant variations and complexities of video temporal sequence \cite{qiu2017learning}. Different from still image analysis, action representations often equip with specific spatial patterns as well as long-term temporal structure. Temporal modeling is critical aspects for action recognition and actions can be characterized by the temporal evolution of appearance governed by motion. Thus, it is crucial to design model which has the capacity to exploit long-range temporal information.

Most recent works for action recognition can be generally originated from three kinds of architectures or frameworks, namely (1) 2D ConvNets with temporal modeling on top, like LSTM \cite{donahue2015long}, (2) 3D based spatiotemporal convolutions \cite{tran2015learning} \cite{ji20133d}, (3) Two-stream based architectures \cite{simonyan2014two} \cite{feichtenhofer2016convolutional} \cite{wang2016temporal}. Long term temporal modeling encode temporal relationship on frame-level features but has a poor capacity of capturing finer temporal relationship. Limited by complex spatiotemporal dependencies of action and computational cost, 3D based ConvNets have been so far hard to scale in terms of recognition performance. Whereas, two-stream based ConvNets \cite{simonyan2014two} which consists of motion and appearance streams typically train separately for each stream, and fuse the outputs in the end. Two-stream based ConvNets have been shown to outperform the 3D based convolution and 2D ConvNets with temporal modeling because they can easily utilize the pre-trained deep architectures \cite{he2016deep} for still-image recognition and have excellent motion sources to extract features.

Nevertheless, some motion features of different class extracted from two-stream framework are prone to confusing, resulting in the wrong classification, due to the similarity structure of optical flow field, for example, discriminating ``eat'' and ``drink'' from ``smoking'' (see Figure \ref{fig:example}). What's more, simple fusing the clip scores of RGB ConvNet and flow ConvNet don't give large improvement. Experiments prove that existing two-stream based frameworks usually failed on categorizing those easily confused action label. Figure \ref{fig:example} give an example, experimental data stemmed from baseline model TSN \cite{wang2016temporal} on HMDB-51 \cite{kuehne2013hmdb51}, of that case. Our human being can easily distinguish above action partly due to we focus on not only the motion features but also the appearance features when we determine an action. Hence, the reason for this case may be two stream based action recognition methods extract spatial and motion features separately, suffering from a limitation of lack of mutual spatial-temporal learning. An excellent framework should be able to capture both information simultaneously. The RGB frames ConvNets should help optical flow ConvNets in features extraction. That is to say, the features learned by the two distinct networks should enhance each other to make the features of the same class compact whereas the different class dissimilar. Actually, there is some subtle connection that is not well explored between spatial ConvNets and temporal ConvNets.

\begin{figure*}[t]
    \centering
    \includegraphics[width=0.99\textwidth]{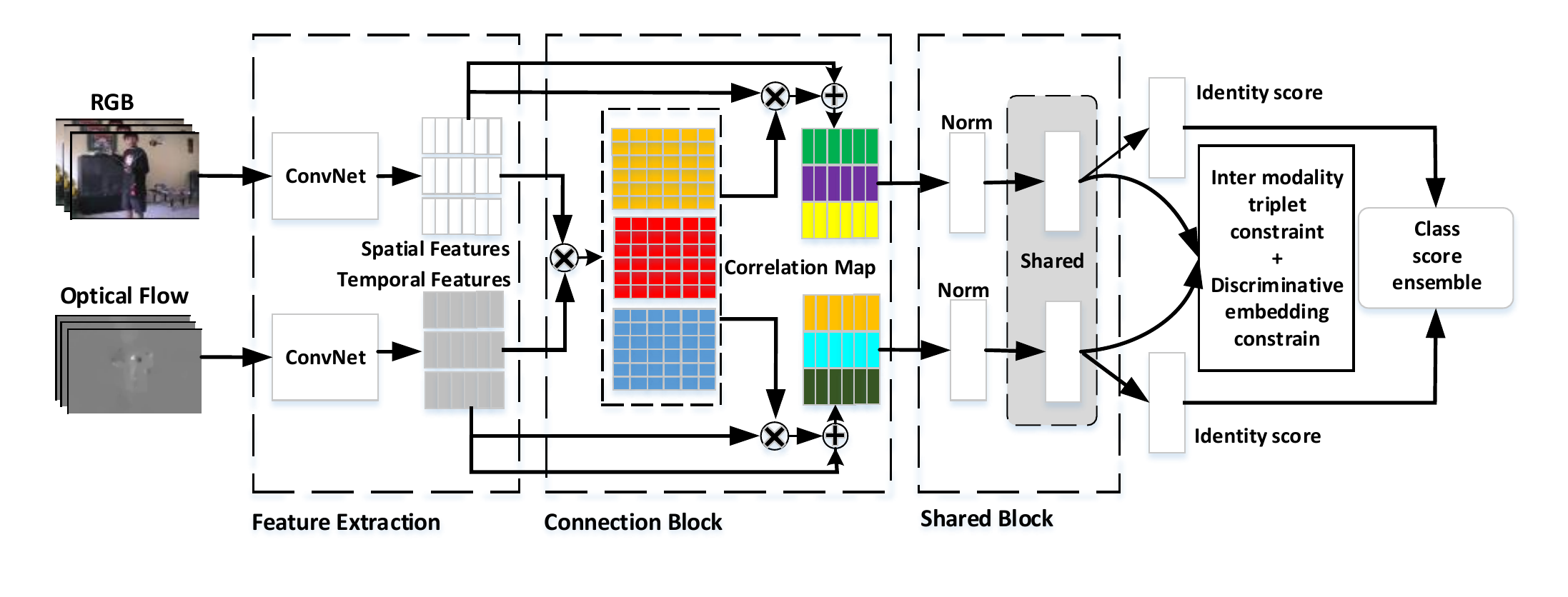}
    \caption{The overall architecture of our proposed cooperative cross-stream network (CCS). Feature extraction ConvNet, connection block, and shared block constitute our model. The feature extraction ConvNet is applied to capturing appearance and motion features. The connection block is used for enhancing appearance and motion features interaction. The shared block is designed for reducing the undesired modality discrepancy. The hole model is training under inter modality triplet and discriminative embedding constraint. The class scores of all modalities are then fused for prediction.}
    \label{fig:model}
\end{figure*}

We introduce an architecture outline in which we simultaneously extract discriminative features and jointly train spatial-temporal network in an end-to-end manner for solving this issue. To efficient explore the relation of the RGB stream and optical flow stream, We propose a cross-modality features extraction paradigm to jointly learning spatiotemporal features for two heterogenous modalities, integrating modality information complementarity block and cross-modality ranking constraint to bridge the gap between two modalities and enhance the modality-invariance of the learned representation. As illustrated in Figure \ref{fig:model}, the main of the framework is spatial and temporal features learning and interaction of the separate part. ConvNet operates the spatial and temporal feature extraction. Inspired by the non local block that calculates the dependency of the same modality features, here we design a new block which takes the spatial and temporal features as input and calculate its' correlation. The connection block which design to capturing dependencies and relationship of spatial and temporal features tries to enhance the interaction between spatial and temporal ConvNets and provide complementarity information to each other. To fully utilize the complementarity information of spatial and temporal features, in the shared block of our framework, we propose to use triplet constraint to force the spatial and temporal features to preserve the similarity structure and weaken the modality discrepancy.

The key contributions of our work are summarized as three-fold: (1) We propose a cooperative spatial and temporal features learning model in an end-to-end manner. Comparing to exist two-stream networks, our model is uniquely able to cope with the incoordination problem between spatial and temporal features extracted in a separate manner. (2) The proposed network enhance the interaction and correlation between the spatial and temporal features by pulling a connection block between the spatial and temporal stream. (3) we aggregate the identity loss with cross-modality ranking constraint to ensure the discriminability by exploiting the relation between spatial and temporal stream.

\section{Related work}
\label{S:2}

As one of video analysis task, action recognition has been well studied for decades. Action recognition is hard partly due to the large inter-class similarity of different action temporal features and intra-class variability of same action spatial features. In this paper, we apply jointly spatial and temporal features learning in a discriminative fashion to improve connections between the two, which cloud in some way learn compact features. Many previous works related to this problem fall into two categories in term of feature learning: (1) hand-crafted features designing, and (2) ConvNets for auto-features extraction.

\noindent\textbf{Hand-crafted features for action recognition.} Before deep learning became popular, most of the traditional CV algorithm variants apply shallow hand-crafted features to solve action recognition. Improved Dense Trajectories (IDT) \cite{wang2013action} which uses densely sampled trajectory features indicates that the temporal information could be processed differently from that of spatial information. Instead of extending the Harris corner detector into 3D, it utilizes the warp optical flow field to obtain some trajectories and eliminate the effects of camera motion in the video sequence. For each tracker corner hand-crafted features, like HOF, HOG, and MBH, are extracted along the trajectory. Despite their excellent performance, IDT and its improvements \cite{peng2016bag}, \cite{pan2016fast}, \cite{wang2015action} are still computationally formidable and become intractable on large-scale datasets.

\noindent\textbf{ConvNets for auto-feature extraction.} An activate research which devotes to the design of deep networks for video representation learning has been trying to devise effective ConvNet architectures \cite{karpathy2014large} \cite{varol2018long} \cite{tran2017convnet} \cite{varol2018long} \cite{donahue2015long}. Karparthy et al. \cite{karpathy2014large} attempt to design a deep network which stacks CNN-based frame-level features in a fixed size and then conduct spatiotemporal convolutions for video-level features learning. However, the results which implied the difficulty of CNNs in capturing motion information of the video is not satisfied. Later, many works in this genre leverage ConvNets trained on frames to extract low-level features an then perform high-level temporal integration of those features using pooling \cite{wang2018video} \cite{wang2018learning}, high-dimensional feature encoding \cite{Girdhar2017ActionVLAD} \cite{diba2017deep}, or recurrent neural networks \cite{donahue2015long} \cite{wu2015modeling} \cite{varol2018long} \cite{yue2015beyond}. Recently, the CNN-LSTM frameworks \cite{donahue2015long} \cite{wu2015modeling}, using stacked LSTM network to connect frame-level representation and exploring long-term temporal relationships of video for learning a more robust representation,  have yielded an improvement for modeling temporal dynamics of convolution features in videos. However, this genre using CNN as an encoder and RNN as a decoder of the video will lose low-level temporal context which is essential for action recognition.

These works implied the importance of temporal information for action recognition and the incapability of CNNs to capture such information. To exploiting the temporal information, some studies resort to the use of the 3D convolution kernel. Tran et al. \cite{tran2015learning} \cite{tran2017convnet} apply 3D CNN, both appearance and motion features learned with 3D convolution, simultaneously encode spatial and temporal cues. Several works explored the effect of performing 3D convolutions over the long-range temporal structure with ConvNets \cite{wang2018two} \cite{yao2015describing}. Unfortunately, the network accepts a predefined number of frames as the input, and it's unclear of the right choice of the temporal span. What's more, the 3D convolution kernel inevitably has more network parameters. Therefore, recent interests have proposed a variant of factorizing a 3D filter into a combination of a 2D and 1D filter, including ``R(2+1)D'' \cite{tran2018closer}, ``Pseudo3D network'' \cite{qiu2017learning}, ``factorized spatiotemporal convolutional networks'' \cite{sun2015human}.

Another efficient way to extract temporal features is to precomputing the optical flow \cite{sun2018optical} using traditional optical flow estimation methods and training a separate CNN to encode the precomputed optical flow, which is kind of escape from temporal modeling but effective in motion features extraction. The famous two-stream architecture \cite{simonyan2014two} proposed to apply two CNN architectures separately on visual frames and staked optical flows to extract spatiotemporal features and then fuse classification score. Further improvements base on this architecture including multi-granular structure \cite{song2019temporal} \cite{zhu2018end}, convolutional fusion \cite{feichtenhofer2016convolutional} \cite{wang2018two}, key-volume mining \cite{zhu2016key}, temporal segment networks \cite{wang2016temporal} and ActionVLAD \cite{Girdhar2017ActionVLAD} for video representation learning. Remarkably, a recent work (I3D) \cite{carreira2017quo} which combines two-stream processing and 3D convolutions holds the state-of-art action recognition results. The work reflects the power of ultra-deep architectures and pre-trained models.

Two-stream architectures based methods generally have the best performance among those works. Nevertheless, two-stream backbone networks often train spatial and temporal ConvNet separately, which will break the connections between appearance and motion information. Recently, many works have utilized cross-modality learning which could improve the discriminative of features to tackle computer vision task, like image retrieve \cite{carvalho2018cross, shen2017deep, shen2015supervised} and person Re-ID \cite{xu2017learning}. In our framework, we jointly train spatial and temporal stream by cross-stream learning. Besides, to capture the complementarity information between appearance and motion across videos and encode the correlation features between different stream into a compact format, we propose a connection block and aggregate inter-modality triplet and intra-modality discriminative embedding constraint with identity loss.

\section{Proposed Method}
\label{S:3}

In this section, we illustrate the framework of our proposed architectures showed in Figure \ref{fig:model}. In our cross-stream network, the spatial stream focuses on appearance features learning from sparsely sample frames, and the temporal stream focus on the motion features which is captured using multiple optical flows. The two parts should complementary to each other; a connection block is designed for improving the interaction of the two different modality features. The latter cross-modality feature learning focuses on learning a multi-modality sharable space to bridge the gap between two heterogenous modalities.

\subsection{Feature Extraction}

We adopt the off-the-shelf features extractor to extract the features from two heterogenous modalities. Both spatial and temporal ConvNet employ similar backbone structures in our feature extraction block.

Suppose we have a video $V_{i}$ containing $T_{i}$ frames,  equipped with a label $l_{i}$, where $l_{i} \in {1, 2, 3, ..., n}$,  $n$ is the total number of action labels. Considering the video $V_{i}$, firstly, we need to get snippet-level action features. A end-to-end deep neural network perform effective video-level representation learning. Here, we use two-stream based framework \cite{simonyan2014two} to extract appearance and motion feature. Given the input $x_{i}^{t}=(s_{t}, F_{t})$, where $s_{t}$ is the $t-th$ frame in video $X_{i}$, $F_{t}=\left\{f_{t^{\prime}}\right\}_{t^{\prime}=t-c1}^{t+c2}$ is stacked optical flow field derived around $s_{t}$,  $c1$, $c2$ are constant, typically 5 $x-level$ images and 5 $y-level$ images. Two-stream network includes spatial and temporal networks which operate on single video frame $s_{t}$ and stacked optical flow field $F_{t}$ respectively. considering the output of $x_{i}^{t}$, $o_{i}^{t}=(x_{i,t}^{f}, x_{i,t}^{o})$, where $x_{i,t}^{f}$ is the learned features of $t-th$ frame in the $i-th$ video, and $x_{i,t}^{o}$ is the learned features of the stacked optical flow $F_{t}$ in the $i-th$ video.

\subsection{Connection Block}

Considering the output features of ConvNet from a video, the features of the sequence $k$ will be $o^{k}$. Here, we suppose the output feature sequences with the appearance and motion have the same size, i.e., $x^{f}, x^{o} \in\mathbb{R}^{ T\times C \times D }$, where $T$, $C$, $D$, denote the sequence number, filter number, and numbers of the output feature dimension. The goal of interaction block is to produce a vector which represents the correlation between $x_{k}^{f}$ and $x_{k}^{o}$, which can be further fed into a neural network to compute the similarity.

Inspired by non-local operation for capturing long-range dependencies \cite{wang2018non} and relationship reasoning module \cite{santoro2017simple} and video temporal reasoning \cite{zhou2018temporal}, we present a pairwise spatial and temporal correlation function as blow:
\begin{equation}
    \mathbf{Y} = \sum_{i, j} g_{\theta}\left(x_{i}^{f}, x_{j}^{o}\right),
\end{equation}
where the input is a set of feature sequences of standard CNN extracted from video frames and optical flows, $x_{i}^{f}$ is the frame feature sequences of $i-th$ video and $x_{j}^{o}$ is the optical flow feature sequence of $j-th$ video, and  $g_{\theta}$ are function typically implement by multiple layer perceptrons with parameter $\theta$ respectively.

Following non local module \cite{wang2018non} aiming at calculating relation of elements of the object, here, we adopt embedded Gaussian to compute the similarity of two different modality object pairs.  Considering the similarity measure function $g_{\theta}$, we present it as follow:
\begin{equation}
    g_{\theta}(x_{i}^{f}, x_{j}^{o}) = e^{\varphi(x_{i}^{f})^{T} \kappa(x_{j}^{o})},
\end{equation}
where $ \theta = \{\varphi, \kappa \}$, and $ \varphi(x_{i}^{f})=W_{\varphi}x_{i}^{f}, \kappa(x_{j}^{o})=W_{\kappa}x_{j}^{o} $ are two embeddings implemented by multiple layer perceptrons.

\begin{figure}[h]
    \centering
    \includegraphics[width=0.7\columnwidth]{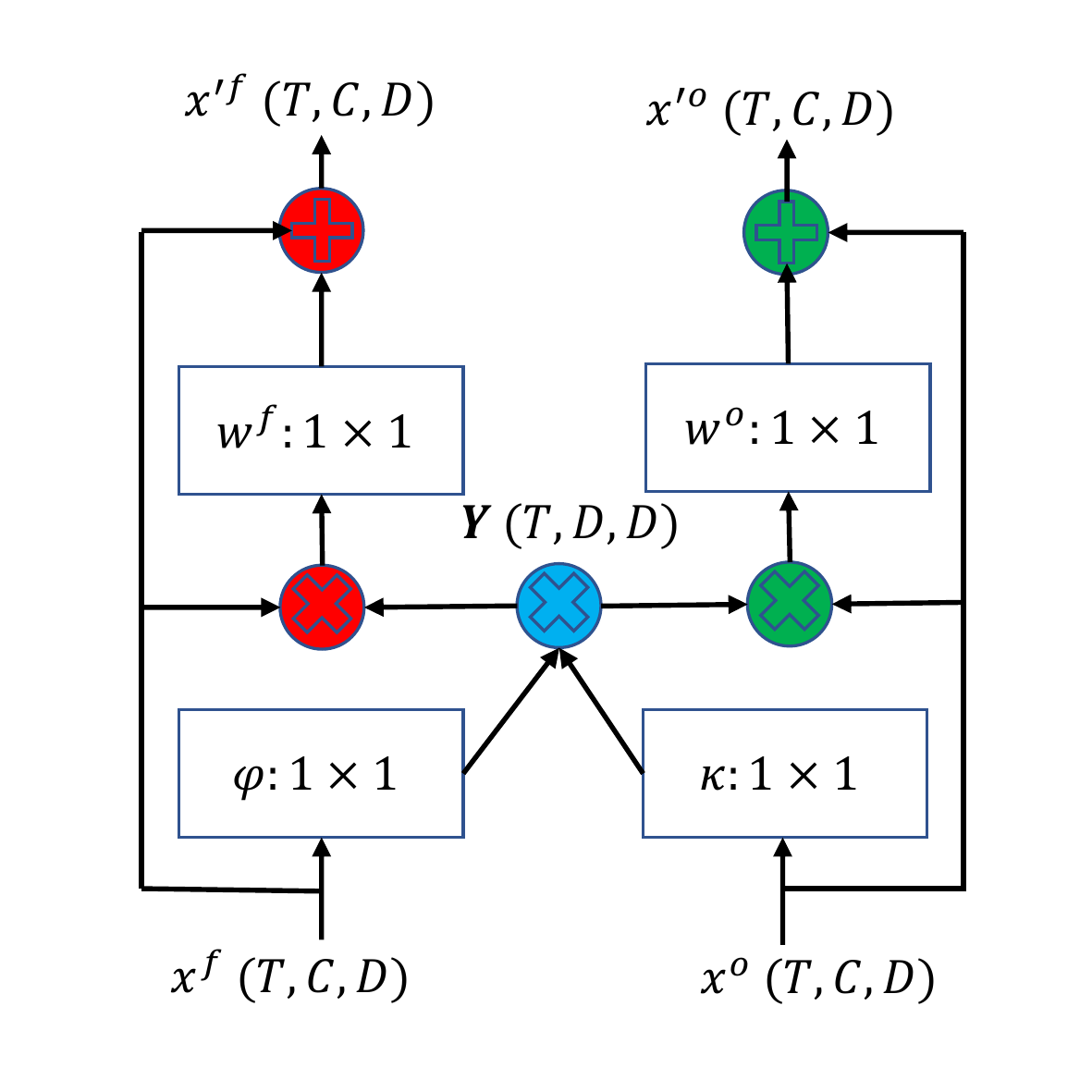}
    \caption{Cross-stream connection block. The appearance features denoted as $x^{f} $ and motion features denoted as $x^{o}$ are put into the connection block. ``$\otimes$'' denotes matrix multiplication, and ``$\oplus$'' denotes element-wise sum. $\varphi$, $\kappa$, $W^{f}$ and $W^{o}$ denote $1 \times 1$ convolution operation.}
    \label{fig:connectin}
\end{figure}

We further wrap the spatial and temporal correlation reasoning Eq.(1) into interaction operation as:
\begin{equation}
      \begin{array}{l} x_{i}^{\prime f} = h_{\phi}^{f}\left(\mathbf{Y}\right) + x_{i}^{f} \\
 x_{j}^{\prime o} = h_{\phi}^{o}\left(\mathbf{Y}\right) + x_{j}^{o}, \end{array}
\end{equation}
where $\mathbf{Y}$ is given in Eq. (1) and $h_{\phi}^{f}\left(\mathbf{Y}\right) = w^{f}x_{i}^{f}y$, $h_{\phi}^{o}\left(\mathbf{Y}\right) = w^{o}x_{j}^{o}y$ are interaction function implemented by a convolution operation. Figure \ref{fig:connectin} shows the details of connection block.

\subsection{Shared Block}

Based on the output features of the ConvNets for $k-th$ segment from a video, it can be transformed into a $D-dimension$ features vector through aggregation operation. Supposing that there is a collection of $n$ instance video features, denoted as $O = \left\{ o _ { i } ^{\prime} \right\} _ { i = 1 } ^ { n }$, $o _ { i } ^{\prime} = \left( z _ { i } ^ { f } , z _ { i } ^ { o } \right)$, where $z _ { i } ^ { f }$ is the aggregation feature of frame-stream ConvNet of $x_{i}^{\prime f}$ and $z_{i}^{ o}$ is the aggregation feature of optical flow filed stream ConvNet of $x_{i}^{\prime o}$, we build learning scheme by selecting triplets from above databases.

Inspired by triplet loss to learn discriminative embedding, we propose to use triplet constraint to extract spatial and motion features based on two-stream backbone ConvNets. Besides, to efficient explore the relation of the RGB stream and optical flow field stream, we propose cross-modality features extraction to jointly learning spatial-temporal features. Most works train the spatial ConvNet and temporal ConvNet separately under the architecture of two-stream. Actually, the RGB frame ConvNet should help optical flow ConvNet in features extraction. That is to say, the features learned by the two distinct networks should enhance each other to make the features of the same class compact whereas the different class dissimilar. The underlying idea is that we compare the distance of a positive appearance-motion pair and the minimum distance of all related negative appearance-motion pairs, rather than each of the negative pairs. More specifically, we sample frames from the entire video and extract appearance and motion features jointly using cross-modality training to enhance the connections of appearance and motion. The extracted features are then fed into a classifier which outputs the classification scores. Final results are improved by scores fusion.

\begin{figure}[h]
    \centering
    \includegraphics[width=0.8\columnwidth]{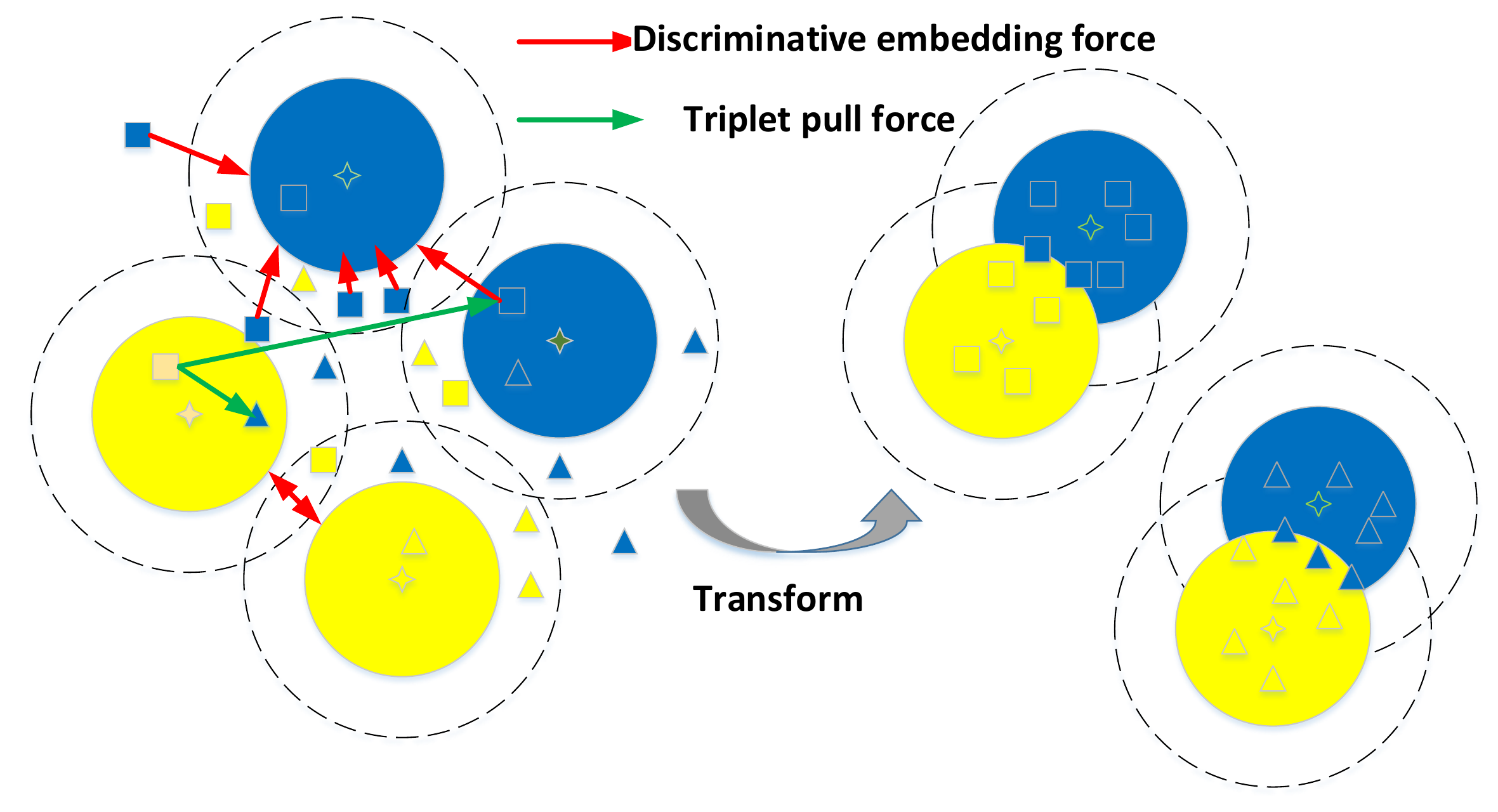}
    \caption{Illustration of the joint effect of inter-modality triplet and discriminative embedding constraint. Different color represents different modality while the same color indicates the class-related cross-modality item; what's more, different shape represent the different class.}
    \label{fig:triplet}
\end{figure}

We propose a cross-modality learning scheme relied on selecting triplets and discriminative embedding scheme on each modality in this section to reduce variations in both intra-modality and cross-modality. Online triplet sampling on each mini-batch \cite{hermans2017defense} are employed here. The joint effect of these two processes is illustrated in Figure \ref{fig:triplet}. The discriminative embedding loss force the learned features of the same class compact and different class dissimilar, meanwhile, the cross-modality triplet loss force appearance and motion stream to project into common feature space.

A set of triplet samples $\left\{z_{a, i}^{f}, z_{p, i}^{o}, z_{n, i}^{o} \right\}_{i}$ and $\left\{z_{a. i}^{o}, z_{p, i}^{f}, z_{n, i}^{f}\right\}_{i}$ are built here, where $z_{r, i}^{m}$ is the member of the triplet, $m \in \{f, o\}$ denote modality, $r \in \{a, p, n \}$ denote the kinds of match of the triplet. The inter-modality loss function using the following expression:
\begin{equation}
\begin{split} 
    \mathcal{L}_ { 1 } = & \sum _ { i } \left( \left[ \left\| z _ { a , i } ^ { f } - z _ { p , i } ^ { o } \right\| _ { 2 } ^ { 2 } - \left\| z _ { a , i } ^ { f } -  z _ { n , i } ^ { o }\right\| _ { 2 } ^ { 2 } + \alpha _ { 1 } \right] _ { + } \right) + \\
    & \sum_{i} \left( \left[ \left\| z _ { a , i } ^ { o } - z _ { p , i } ^ { f } \right\| _ { 2 } ^ { 2 } - \left\| z _ { a , i } ^ { o } - z _ { n , i } ^ { f } \right\| _ { 2 } ^ { 2 } + \alpha _ { 1 } \right] _ { + } \right), 
\end{split}
\end{equation}
where $\alpha_{1}$ is a margin that is enforced between positive and negative pairs.

Since the above ranking loss constrains the feature learning process with their underlying relationships among the heterogeneous modality, it's hard to learn a robust feature representation to reduce the intra-class variations by simply exploiting the relationship cues. Inspired by linear discriminative analysis \cite{Belhumeur2002Eigenfaces}, we introduce discriminative embedding constraint to enhance the robustness of the learned feature representation and address intra-modality variations; the discriminative embedding loss function expresses as following:
\begin{small}
\begin{equation}
\begin{split}
    \mathcal{L} _ { 2 } \!= & \!\left( \sum _ { i , c_{i} } \left[ \left\| z _ { i }^{f} \!-\! m _ { c_{i} }^{f} \right\| _ { 2 } ^ { 2 } \!-\! \alpha _ { 2 } \right] _ { + } \!+ \!\sum _ { c _ { i } \neq c _ { j } } \left[ \alpha _ { 3 } \!-\! \left\| m _ { c _ { i } }^{f} \!-\! m _ { c _ { j } }^{f} \right\| _ { 2 } ^ { 2 } \right] _ { + } \right) \!+\! \\    
    & \left(\sum _ { i , c_{i} } \left[ \left\| z _ { i }^{o} \!-\! m _ { c_{i} }^{o} \right\| _ { 2 } ^ { 2 } \!-\! \alpha _ { 2 } \right] _ { + } \!+\! \sum _ { c _ { i } \neq c _ { j } } \left[ \alpha _ { 3 } \!-\! \left\| m _ { c _ { i } }^{o} \!-\! m _ { c _ { j } }^{o} \right\| _ { 2 } ^ { 2 } \right] _ { + } \right),
\end{split}
\end{equation}
\end{small}
where $m _ { c _ { i } }$ is the mean feature of class $i$, $c$ is the number of the class and $\alpha_{2}$ is a margin that forces the same class compact, $\alpha_{3}$ is a margin that is enforced between the different class.

For the sake of feasibility and effectiveness in classification, the general cross entropy loss $\mathcal{L}_{3}$ is utilized by treating each action as a class. In this manner, the identity-specific information is integrated to enhance the robustness.

Based on the above, the loss function of the proposed network, referred to as a combination of cross-modality, is formulated as the combination of the intra-modal discrimination loss and the intra-modal embedding constraint  and identity loss:
\begin{equation}
    \mathcal{L} = \lambda_{1} \mathcal{L}_{1} + \lambda_{2} \mathcal{L}_{2} + \mathcal{L}_{3},
\end{equation}
where $\lambda_{1}$, $\lambda_{2}$ control the contribution of the two terms. Algorithm \ref{algo:1} illustrates the steps of the proposed cooperative cross-stream network. From the backward pass, we can obtain that the connection block crosses the stream function as a bridge for information of appearance and motion stream flowing to each other.

\begin{algorithm}[h]
    \caption{Optimization step of CCS.}
    \begin{algorithmic}
    \STATE \textbf{Input:}~ $N$ videos with $n$ class $\left \{ (X_{i}, l_{i}) \right \}_{i=1}^{N}$, where $l_{i} \in \left \{1, 2, \ldots, n \right \}$ is the label of video $X_{i}$, iteration number $K$.
    \STATE \textbf{Output:}~ The predicted action label $Y=\left\{y_{i}\right\}_{i=1}^{N}$, where $y_{i} \in \left\{1, 2, \ldots, n\right\}$ \\
    \STATE \textbf{initialization:} $i=0$,\\
    \STATE \textbf{repeat}\\
\quad \quad 1. \textbf{Forward pass}: \\
\quad \quad \quad 1.1 compute the appearance features $x_{i}^{\prime f}$ and motion \\
\quad \quad \quad   features $x_{i}^{\prime o}$ within the connection block;\\
\quad \quad \quad 1.2 predict the video label $y_{i}$ after shared block;\\
\quad \quad 2. \textbf{Backward pass}:\\
\quad \quad \quad using $ \frac{\partial \mathcal{L}}{\partial \theta_{f}} = \frac{\partial \mathcal{L}}{\partial z^{f}}\frac{\partial z^{f}}{\partial \theta^{f}} + \frac{\partial \mathcal{L}}{\partial z^{o}}\frac{\partial z^{o}}{\partial \theta^{f}}$ and\\
\quad \quad \quad $ \frac{\partial \mathcal{L}}{\partial \theta_{o}} = \frac{\partial \mathcal{L}}{\partial z^{f}}\frac{\partial z^{f}}{\partial \theta^{o}} + \frac{\partial \mathcal{L}}{\partial z^{o}}\frac{\partial z^{o}}{\partial \theta^{o}}$ as parameters gradient,\\
\quad \quad \quad where $\theta_{f}$ are the parameters of spatial stream model and \\
\quad \quad \quad $\theta_{o}$ are the parameters of temporal stream model;\\
\quad \quad 3. $i = i + 1$;\\
\STATE \textbf{until} $i = K$ or convergence
\end{algorithmic}
\label{algo:1}
\end{algorithm}

\section{Experiments}
\label{S:4}

In this section, we describe our method for action recognition. Firstly, we introduce the benchmark datasets and implementation details of the proposed method. Afterward, we compare our method with state-of-art methods on standard action datasets. Following, we explore the effectiveness of applying different component in our proposed model. Finally, we investigate the effect of ConvNet architectures and hyperparameters and visualize the interesting region extracted by our model on the snippet video.

\subsection{Experimental Setup}

\noindent\textbf{Datasets.} We conduct our experiments on three challenging action datasets: namely, UCF-101 \cite{soomro2012ucf101}, HMD-B51 \cite{kuehne2013hmdb51}, something-something-V2 \cite{mahdisoltani2018fine} to evaluate the overall performance. The UCF-101, one of popular action recognition dataset, consists of 101 action classes with 13320 short video clips. Videos in this dataset have $320 \times 240$ spatial resolution. The HMDB-51 dataset has 6766 video clips with 51 categories. something-something-v2 an interesting temporal relationship reasoning dataset contain total 220,847 video clips with 174 action classes. For both datasets, we follow the standard evaluation protocol and adopt its training/testing splits for evaluation. We report accuracy on the split 1 test set of UCF-101 and HMDB-51 datasets.

\noindent\textbf{Implementation details.} We employ the pytorch framework in this paper for Networks building, and all networks are trained on two GeForce GTX Titan X GPU with total 24G memory. We compute optical flow with a TV-L1 algorithm \cite{perez2013tv}. All the input images are resized to 224 $\times$ 224 followed by the dataset processing strategy of \cite{wang2016temporal}. We adopt mini-batch stochastic gradient descent optimizer for model training, and initial learning rate here is 0.001 which will reduce by a factor 10 after 50 epochs. It has decay rate $5 \times 10^{-4}$, and momentum 0.9 to update Network parameters. The maximum epochs are 400. The trade-off parameters $\lambda_{1}$ and $\lambda_{2}$ are all set as $0.5$. We set cross-modality margin $\alpha_{1}=0.3$, and intra-class $\alpha_{2}=0.3$, inter-class margin $\alpha_{3}=0.8$ of the same modality.

We introduce a balance mini-batch sampling strategy for inter-modality modality constraint and discriminative embedding constraint. Specifically, we randomly select $N$ action categories. Then we randomly select $M$ instance of the selected identity from two different modalities to construct the mini-batch, in which totally $2 \times N \times M $ instances are fed into the network for training.

\subsection{Comparison with Existing Methods}
We compare to the state-of-the-art action recognition methods and report the results in Table \ref{tab:compare} on UCF-101, HMDB-51 (split 1) and something-something-V2 dataset. For a fair comparison, we list the important factor such as the pre-trained dataset and use RGB images and optical flow fields as input modalities. We use CCS, as above, and predict the action in a single forward pass using fully network testing. Here, we extract three segments of a video and randomly sample a video snippet of 10 frames on each segment as input for training. During testing, 25 frames are sampled for each video. The comparison against the single model without ensemble technique, like the work in \cite{donahue2015long}, which attaches an LSTM to a ConvNet architecture and the one spatiotemporal C3D based network \cite{tran2017convnet} are impressive. Their accuracy of $85.8\%$ is to date the best performing approach using one stream for action recognition. Here, our gain of $12.3\%$ further underlines the importance of two-stream framework. Comparing to the original two-stream method \cite{simonyan2014two}, we improve by $9.7\%$ on UCF-101 and by $22.9 \% $on HMDB-51. Apparently, even though the original two-stream approach has the advantage than one stream method, the benefit of our cooperative cross-stream network with the interaction of heterogeneous features are still greater. Together with TSN or I3D, our cooperate two-stream architecture widens the advantage over previous models considerably, bringing overall performance to $97.4\%$ on UCF-101 and $81.9\%$ on HMDB-51. We observe that the combination of RGB images and optical flow image boosts the recognition performance and cooperative training the two kinds of image further yield an improvement. This result indicates that RGB images and optical flow image may encode complementary information.

These relatively larger performance increments again underline that our approach is better able to capture the available dynamic information. Overall, our result $81.9\%$ on HMDB-51 clearly sets a new state-of-the-art on this widely used action recognition datasets. This corroborates for different modality information, enhanced by modality connection block and cross-modality training, is crucial for a better understanding of action in videos. What's more, from Table \ref{tab:compare}, we also can acquire the power of pre-trained model for action recognition.

something-something-v2 is a dataset for human-object interaction recognition, which cares more about temporal relations and transformations of objects rather than the appearance and motion of the objects characterize the activities \cite{zhou2018temporal}. In Table \ref{tab:something}, We report the accuracies of something-something-v2. Comparing with the baseline methods \cite{goyal2017something}, our method further improves to $61.1\%$. The combination of two-stream TRN \cite{zhou2018temporal} and our CCS achieves better results. The performance demonstrates the importance of not only the temporal reasoning pooling but also the correlation of appearance and motion features on something-something dataset.

\begin{table*}
    \caption{Comparison of state-of-the-art methods on the UCF-101 and HMDB-51 datasets (split 1). We report the accuracy of RGB modality, optical flow modality, and the combination of both two modality.}
   \vspace{-3mm}
      \begin{center}
      \resizebox{\textwidth}{!}{
      \begin{tabular}{lccccccc}
         \toprule
         \multirow{2}{*}{Methods}&
         \multirow{2}{*}{Pre-train dataset}&
         \multicolumn{3}{c}{UCF-101~}&
         \multicolumn{3}{c}{HMDB-51~}\\
         &&RGB&Flow&RGB+Flow& RGB&Flow&RGB+Flow\\
         \midrule
         ConvNets+LSTM \cite{donahue2015long}~&ImageNet& 68.2 & - & - &- &- & -\\
         Two-stream Network \cite{simonyan2014two}~&ImageNet&73.0 & 83.7 & 88.0  &40.5 &54.6 & 59.4\\
         ConvNet fusion \cite{feichtenhofer2016convolutional}~&ImageNet&82.6&86.2.7&90.6  &47.0 &55.2 & 58.2\\
         ST-resNet \cite{feichtenhofer2016spatiotemporal}~&ImageNet&82.3&79.1& 93.4  &43.2 &55.5 & 66.4\\
         DTPP \cite{zhu2018end}~&ImageNet&89.7&89.1& 94.9 &61.5 &66.3 & 75.0\\
         TLE+Two-stream  \cite{diba2017deep}~&ImageNet& - & -  & 95.6&- &- & 71.1\\
         ActionVLAD \cite{Girdhar2017ActionVLAD}~&ImageNet&~ -&-&92.7 &49.8 &59.1 & 66.9\\
         C3D \cite{tran2015learning}~&sports-1M&82.3&- & - &51.6 &- & -\\
         C3D \cite{tran2017convnet}~&sports-1M&85.8&- & - &54.9 &- & -\\
         R(2+1)D \cite{tran2018closer}~&sports-1M&93.6&93.3 & 95.0 &66.6 &70.1 & 72.7\\
         TSN \cite{wang2016temporal}~&ImageNet&85.7&87.9 & 93.5 &- &- & 68.5\\
         I3D \cite{carreira2017quo}~&ImageNet&84.5&90.6 & 93.4 &49.8 &61.9 & 66.4\\
         R(2+1)D \cite{tran2018closer}~&ImageNet+Kinetics& \textbf{96.8} & \textbf{95.5} & 97.3 & \textbf{74.5} & \textbf{76.4} & 78.7\\
         TSN \cite{wang2016temporal}~&ImageNet+Kinetics&91.1 & 95.2 & 97.0 &- &- & -\\
         \hline
         \textbf{CCS + TSN} ~&ImageNet & 87.2 & 87.4 & 95.3 & 60.5 & 62.1 & 77.2 \\
         \textbf{CCS + TSN} ~&ImageNet+Kinetics & 94.2 & 95.0 & \textbf{97.4} & 69.4 & 71.2 & \textbf{81.9} \\
         \textbf{CCS + I3D} ~&ImageNet & 86.7 & 87.1 & 93.8 & 60.1 & 62.3 & 68.2\\
         \bottomrule
         \label{tab:compare}
      \end{tabular}
      }
   \end{center}
\end{table*}

\begin{table}
    \caption{Results on something-something-V2.}
   \vspace{-3mm}
   \begin{center}
      \begin{tabular}{lcccc}
         \toprule
         \multirow{2}{*}{Methods}&
         \multicolumn{2}{c}{val}&
         \multicolumn{2}{c}{Test} \\
         &top-1&top-5&top-1&top-5\\
         \midrule
         Baseline & 51.3 & 80.6 & - & - \\
         MultiScale TRN & 48.8 & 77.6 & 50.9 & 79.3\\
         two-stream + TRN & 55.5 & 83.1 & 56.2 & 83.2\\
         \textbf{CCS + two-stream + TRN} & \textbf{61.2} &\textbf{89.3} & \textbf{60.5} & \textbf{87.9}\\
         \bottomrule
      \end{tabular}
   \end{center}
   \label{tab:something}
\end{table}

\subsection{Further Analysis}
\noindent\textbf{Importance of each component of the proposed model.} With all the design choices set, we now apply the cooperative cross-stream network (CCS) to the action recognition with different variants, where the result is illustrated in Table \ref{tab:Ablation}. A component-wise analysis of the components in terms of the recognition accuracies is also presented.

\begin{table*}[t]
    \caption{Ablation studies: Results with different components on the UCF-101~, HMDB-51~ datasets. ``method'' denotes the component we use in our final model. ``CB'': with only connection block. ``CS'': without connection module only using cross-stream training in the shared block.  ``All'': with the connection block and cross-stream training in the shared block.}
   \vspace{-3mm}
   \begin{center}
   \resizebox{\textwidth}{!}{
      \begin{tabular}{ccccccccc}
         \toprule
         \multirow{2}{*}{Base Model}&
         \multirow{2}{*}{Methods}&
         \multicolumn{3}{c}{UCF-101~}&
         \multicolumn{3}{c}{HMDB51~} \\
         & & RGB& Flow & RGB+Flow& RGB& Flow & RGB+Flow\\
         \midrule
         \multirow{4}{*}{TSN} & Baseline~ & 85.7 & \textbf{87.9} & 93.5 & - & - & 68.5\\
         & CB &86.3 & 87.2 & 93.9 & 60.5 & 62.1 & 76.3\\
         & CS  & 84.9 & 85.1 & 91.7 & 54.4 & 61.6 & 67.3\\
         & All  & \textbf{87.2} & 87.4 & \textbf{95.3} & \textbf{61.7} &\textbf{65.1}& \textbf{77.2}\\ \hline
         \multirow{4}{*}{I3D}& Baseline &84.5 &\textbf{90.6} & 93.4 & 49.8 & 61.9 & 66.4\\
         & CB & 86.1 & 86.9 & 92.7 & 53.0 & 56.2 & 67.6 \\
         & CS & 82.4 & 83.1 & 91.8 & 50.9 & 52.3 & 64.7\\
         & All & \textbf{86.7} & 87.1 & \textbf{93.8} & \textbf{60.1} &\textbf{62.3}&\textbf{68.2}\\ 
         \bottomrule
      \end{tabular}
      \label{tab:Ablation}
      }
   \end{center}
\end{table*}

We cooperate the CCS with TSN \cite{wang2016temporal} and I3D \cite{carreira2017quo}, to verify the importance of modality information complementarity. Instead of training spatial and temporal stream separately, CCS jointly train the two stream network to improve the interaction of deep spatiotemporal features so that the model not only captures the co-occurrence also the specific patterns in the features. We keep all the training conditions the same, and vary connection block and loss function used by two models.

We investigate the effectiveness of each component in our proposed model by conducting a series of ablation studies on all three datasets. We treat the TSN \cite{wang2016temporal} and I3D \cite{carreira2017quo} as backbone framework in this section. We first study the effectiveness of our modality features connection modules by replacing the connection module with feature concatenation or average. We first train the TSN and I3D framework with the connection module, named TSN+CM, I3D+CM. Its' RGB ensemble with optical flow accuracy increase by $0.4\%$, on the UCF101 dataset, and $7.8\%$ on the HMDB51, which demonstrates that conducting modality information interaction with connection block helps deep modality features complementarity to enhance the performance. For validating the effectiveness of shared features projection layer following connection module, we remove the shared layer and only employ cross entropy loss. Instead, we directly take the results from TSN or I3D and input them into two-layer feed-forward neural networks mentioned above to obtain the similarity confidence (denoted as TSN+CS I3D+CS). The performance even becomes worse compared with TSN and I3D. However, worked with connection block, our original CCS network can achieve the best results.

We can obtain that the reported baselines typically underperform the proposed model. Both TSN \cite{wang2016temporal} and I3D \cite{carreira2017quo} produce reasonable performances but work with our original design still yield improvement. We speculate this is because the connection block considerably explores the correlation information of heterogeneous modality and therefore, the network is able to store more complementary information for cross-modal feature learning.

We consider the additional parameters introduced by connection and shared block. All of them are contained in two 1$\times$1 convolution and shared full connection operation. The computation is relatively small and worth of the cost, compared to the whole networks and contribution to model performance.

\begin{figure}
    \centering
    \subfloat[Aggregation function]{\includegraphics[width=7cm]{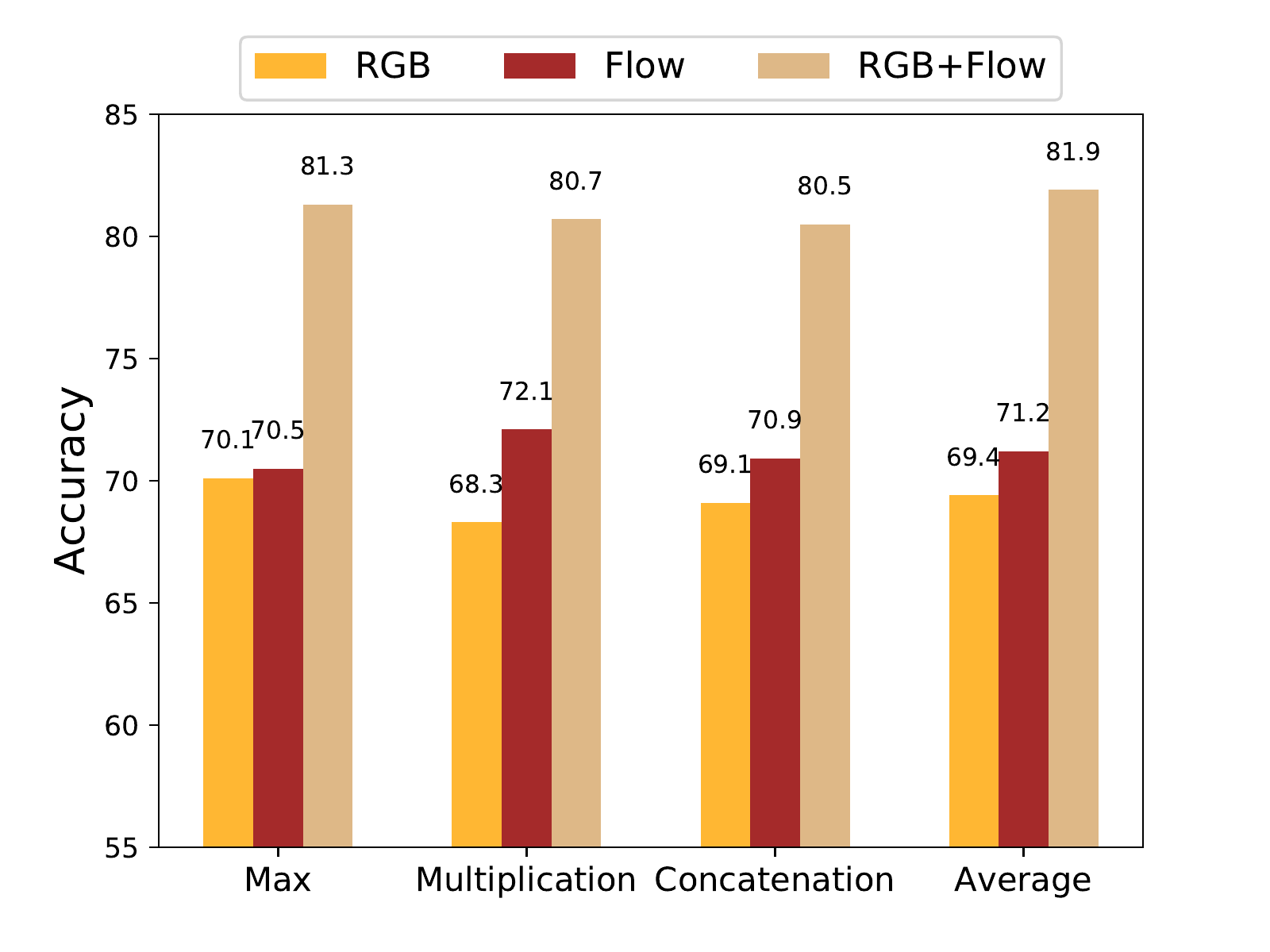}
    }
    \hfil
    \subfloat[Architecture]{\includegraphics[width=7cm]{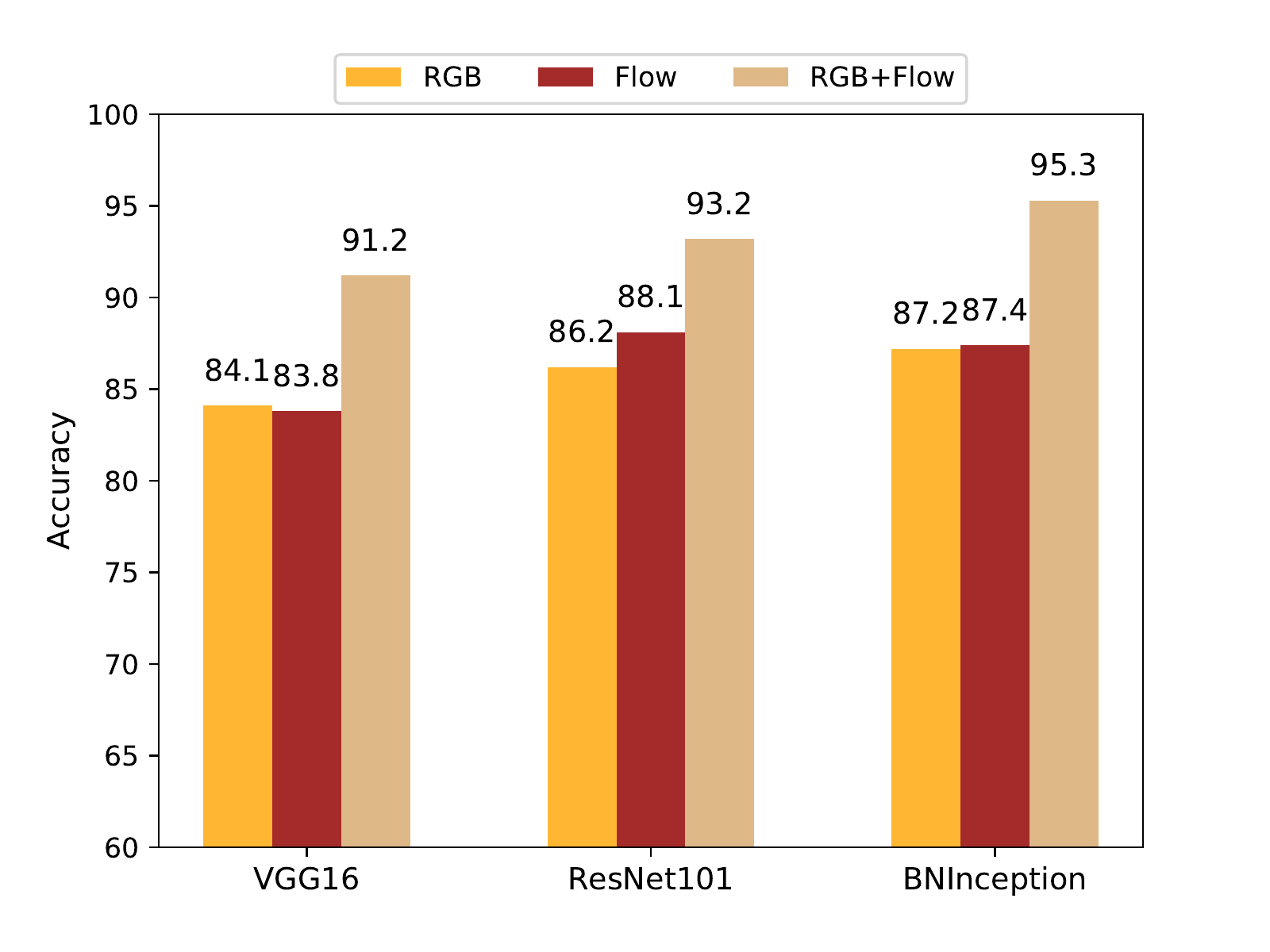}
    }
    \caption{The performance of different backbone architectures or feature aggregation functions.}
    \label{fig:arc_agg}
    \end{figure}

\noindent\textbf{Effect of sequence features aggregation function.} The two commonly used aggregation methods are the element-wise maximum of the sequence and element-wise average of the sequence. Here, we also evaluate (1) element-wise multiplication of the sequence, (2) concatenation of sequence. The comparisons among the four late score fusion methods are shown in Figure \ref{fig:arc_agg} (a). We can see that the element-wise average of the sequence achieves the best result on HMDB-51 dataset. This verifies that the effective of element-wise average to improve the final accuracy.

\noindent\textbf{Effect of model parameter.} We survey the hyperparameter $\alpha_{1}$, $\alpha_{2}$ and $\alpha_{3}$ in ranking loss. The parameter $\alpha_{1}$ refers to the margin between the anchor/positive and negative samples. The parameter $\alpha_{2}$ refers to the margin between the sample of its center and $\alpha_{3}$ refers to the margin between different center. A small value enforces less on the similarities between the anchor/positive against negative, but the loss in faster convergence. On the other hand, a large value may lead to a network with good performance, but slow convergence during training. We conduct an experiment on UCF-101 to illustrate the effects of this parameter, and the results are showed in Table \ref{tab:hyper}. From the table, it can be seen that it achieves the best accuracy when $\alpha_{1}$ and $\alpha_{2}$ are set to $0.3$, and $\alpha_{3}$ is set as $0.8$. This suggests that we should carefully choose the hyperparameter, and it is advisable to set relatively small $\alpha$ values for reasonable results.

\begin{table}
    \caption{The performance of CCS with different model parameters values on UCF-101 dataset.}
   \label{tab:hyper}
   \vspace{-3mm}
   \begin{center}
      \begin{tabular}{cccc}
         \toprule
         \multicolumn{3}{c}{Parameters} &
         \multirow{2}{*}{Accuracy} \\
         $\alpha_{1}$ & $\alpha_{2}$ & $\alpha_{3}$ \\
         \midrule
          0.2 & 0.3 & 0.8 & 96.3  \\
          0.3 & 0.3 & 0.8 & \textbf{97.4}  \\
          0.3 & 0.5 & 1.0 & 97.1  \\
          0.5 & 0.5 & 1.0 & 95.4  \\
          0.8 & 0.5 & 1.2 & 95.5   \\
         \bottomrule
      \end{tabular}
   \end{center}
\end{table}

    
\noindent\textbf{Effect of ConvNet structure. } Furthermore, to investigate different effect of ConvNet structures, We also explore the conventional CNN model, namely VGG \cite{simonyan2014very}, ResNet \cite{he2016deep}, BN-Inception \cite{ioffe2015batch}, all pre-trained on ImageNet, as the backbone of two-stream ConvNets. All those ConvNets are trained together with TSN and our CCS network on UCF-101. The results of those deep structures are shown in Figure \ref{fig:arc_agg} (b). Among those structures, BN-Inception achieves the best accuracy.

\subsection{Visualization}
\begin{figure}[h]
    \centering
    \includegraphics[width=0.99\columnwidth]{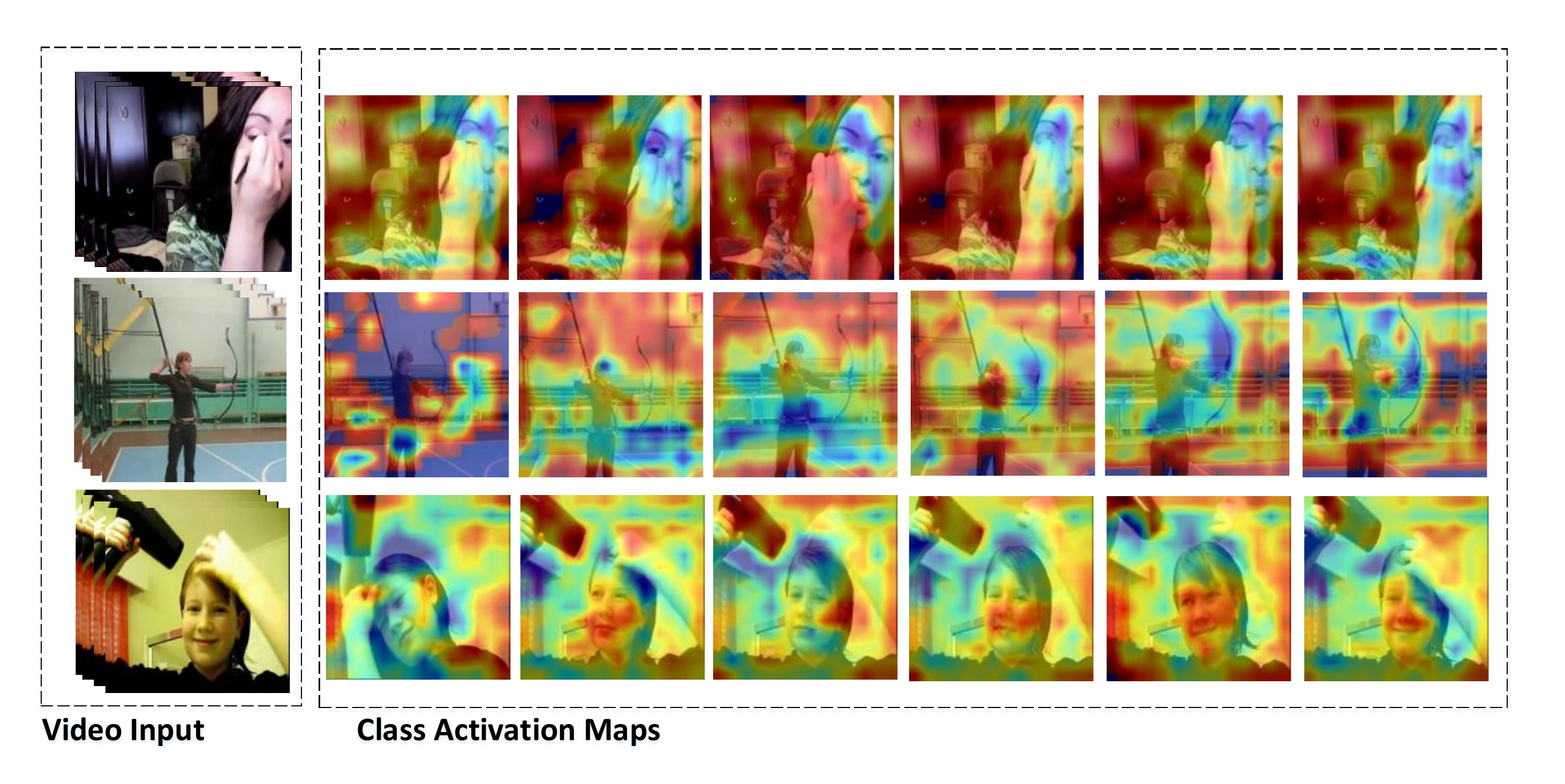}
    \caption{Visualization of ``CAM'' \cite{zhou2016learning} generated by our CCS model when jointly trained appearance and motion stream. The maps highlight the discriminative region for action classification.}
    \label{fig:cam}
\end{figure}

To verify how our model help in action classification, we would like to attain further insight into what our model has learned. As shown in \cite{zhou2016learning},  ConvNets are expert in capturing the basic visual concept, but it has difficulty in identifying the importance of different units for classifying different categories. Here, we use the CAM (Class Activation Map) \cite{zhou2016learning} to visualize the most discriminative parts of the proposed model. Thus the output after a number of iterations can be considered as class visualization based on class knowledge inside the ConvNet model. To understanding the primitives our model used for represent actions and visualizing interesting class information in CCS models, We randomly select three classes from the UCF-101 dataset, ``Apply Eye Make-Up'', ``Archery'', ``Blow Dry Hair'' as visualization example. For ease of visualization, we only consider the spatial stream in this example. The results are shown in Figure \ref{fig:cam}. The highlight regions that correspond to the receptive field give us same insight of what the model cares about. For example, we see that the proposed model pays more attention to the region like `eye' and `hand' in 'ApplyEyeMakeUp' video.

\section{conclusion}
In this paper, a novel CCS network for video action recognition was proposed. It cooperatively exploits the information in RGB visual appearance features and optical flow motion features by mixing a connection block and jointly optimizing a ranking loss and a cross entropy loss. The CCS network enhances the discriminative power and explore the complementary information of the deeply learned heterogeneous features and weakens the modality discrepancy. Further, it can apply to both homogeneous and heterogeneous modality-based action recognition task. The ranking loss consists of inter-modality triplet constraint and discriminative embedding constraint, and it reduces both the intra-modality and cross-modality feature variations. Experiment results on three datasets demonstrate and justify the effectiveness of the proposed method.

\section*{Acknowledgment}
This work was supported in part by the National Natural Science Foundation of China under grants No. 61502081, 61602089, 61632007 and the Sichuan Science and Technology Program 2018GZDZX0032, 2019ZDZX0008 and 2019YFG0003.

\bibliographystyle{ieee}
\bibliography{ccs.bib}

\begin{thebibliography}{10}\itemsep=-1pt

\bibitem{Belhumeur2002Eigenfaces}
P.~N. Belhumeur, J.~P. Hespanha, and D.~J. Kriegman.
\newblock Eigenfaces vs. fisherfaces: Recognition using class specific linear
  projection.
\newblock {\em IEEE Transactions on Pattern Analysis \& Machine Intelligence},
  19(7):711--720, 2002.

\bibitem{carreira2017quo}
J.~Carreira and A.~Zisserman.
\newblock Quo vadis, action recognition? a new model and the kinetics dataset.
\newblock In {\em proceedings of the IEEE Conference on Computer Vision and
  Pattern Recognition}, pages 6299--6308, 2017.

\bibitem{carvalho2018cross}
M.~Carvalho, R.~Cad{\`e}ne, D.~Picard, L.~Soulier, N.~Thome, and M.~Cord.
\newblock Cross-modal retrieval in the cooking context: Learning semantic
  text-image embeddings.
\newblock In {\em The 41st International ACM SIGIR Conference on Research \&
  Development in Information Retrieval}, pages 35--44. ACM, 2018.

\bibitem{diba2017deep}
A.~Diba, V.~Sharma, and L.~Van~Gool.
\newblock Deep temporal linear encoding networks.
\newblock In {\em Proceedings of the IEEE Conference on Computer Vision and
  Pattern Recognition}, volume~1, 2017.

\bibitem{donahue2015long}
J.~Donahue, L.~Anne~Hendricks, S.~Guadarrama, M.~Rohrbach, S.~Venugopalan,
  K.~Saenko, and T.~Darrell.
\newblock Long-term recurrent convolutional networks for visual recognition and
  description.
\newblock In {\em Proceedings of the IEEE conference on computer vision and
  pattern recognition}, pages 2625--2634, 2015.

\bibitem{feichtenhofer2016spatiotemporal}
C.~Feichtenhofer, A.~Pinz, and R.~Wildes.
\newblock Spatiotemporal residual networks for video action recognition.
\newblock In {\em Advances in neural information processing systems}, pages
  3468--3476, 2016.

\bibitem{feichtenhofer2016convolutional}
C.~Feichtenhofer, A.~Pinz, and A.~Zisserman.
\newblock Convolutional two-stream network fusion for video action recognition.
\newblock In {\em Proceedings of the IEEE Conference on Computer Vision and
  Pattern Recognition}, pages 1933--1941, 2016.

\bibitem{Girdhar2017ActionVLAD}
R.~Girdhar, D.~Ramanan, A.~Gupta, J.~Sivic, and B.~Russell.
\newblock Actionvlad: Learning spatio-temporal aggregation for action
  classification.
\newblock In {\em IEEE Conference on Computer Vision \& Pattern Recognition},
  2017.

\bibitem{goyal2017something}
R.~Goyal, S.~E. Kahou, V.~Michalski, J.~Materzynska, S.~Westphal, H.~Kim,
  V.~Haenel, I.~Fruend, P.~Yianilos, M.~Mueller-Freitag, et~al.
\newblock The" something something" video database for learning and evaluating
  visual common sense.
\newblock In {\em ICCV}, volume~2, page~8, 2017.

\bibitem{he2016deep}
K.~He, X.~Zhang, S.~Ren, and J.~Sun.
\newblock Deep residual learning for image recognition.
\newblock In {\em Proceedings of the IEEE conference on computer vision and
  pattern recognition}, pages 770--778, 2016.

\bibitem{hermans2017defense}
A.~Hermans, L.~Beyer, and B.~Leibe.
\newblock In defense of the triplet loss for person re-identification.
\newblock {\em arXiv preprint arXiv:1703.07737}, 2017.

\bibitem{ioffe2015batch}
S.~Ioffe and C.~Szegedy.
\newblock Batch normalization: Accelerating deep network training by reducing
  internal covariate shift.
\newblock {\em arXiv preprint arXiv:1502.03167}, 2015.

\bibitem{ji20133d}
S.~Ji, W.~Xu, M.~Yang, and K.~Yu.
\newblock 3d convolutional neural networks for human action recognition.
\newblock {\em IEEE transactions on pattern analysis and machine intelligence},
  35(1):221--231, 2013.

\bibitem{karpathy2014large}
A.~Karpathy, G.~Toderici, S.~Shetty, T.~Leung, R.~Sukthankar, and L.~Fei-Fei.
\newblock Large-scale video classification with convolutional neural networks.
\newblock In {\em Proceedings of the IEEE conference on Computer Vision and
  Pattern Recognition}, pages 1725--1732, 2014.

\bibitem{kuehne2013hmdb51}
H.~Kuehne, H.~Jhuang, R.~Stiefelhagen, and T.~Serre.
\newblock Hmdb51: A large video database for human motion recognition.
\newblock In {\em High Performance Computing in Science and Engineering ‘12},
  pages 571--582. Springer, 2013.

\bibitem{mahdisoltani2018fine}
F.~Mahdisoltani, G.~Berger, W.~Gharbieh, D.~Fleet, and R.~Memisevic.
\newblock Fine-grained video classification and captioning.
\newblock {\em arXiv preprint arXiv:1804.09235}, 2018.

\bibitem{pan2016fast}
Z.~Pan, P.~Jin, J.~Lei, Y.~Zhang, X.~Sun, and S.~Kwong.
\newblock Fast reference frame selection based on content similarity for low
  complexity hevc encoder.
\newblock {\em Journal of Visual Communication and Image Representation},
  40:516--524, 2016.

\bibitem{peng2016bag}
X.~Peng, L.~Wang, X.~Wang, and Y.~Qiao.
\newblock Bag of visual words and fusion methods for action recognition:
  Comprehensive study and good practice.
\newblock {\em Computer Vision and Image Understanding}, 150:109--125, 2016.

\bibitem{perez2013tv}
J.~S. P{\'e}rez, E.~Meinhardt-Llopis, and G.~Facciolo.
\newblock Tv-l1 optical flow estimation.
\newblock {\em Image Processing On Line}, 2013:137--150, 2013.

\bibitem{qiu2017learning}
Z.~Qiu, T.~Yao, and T.~Mei.
\newblock Learning spatio-temporal representation with pseudo-3d residual
  networks.
\newblock In {\em 2017 IEEE International Conference on Computer Vision
  (ICCV)}, pages 5534--5542. IEEE, 2017.

\bibitem{santoro2017simple}
A.~Santoro, D.~Raposo, D.~G. Barrett, M.~Malinowski, R.~Pascanu, P.~Battaglia,
  and T.~Lillicrap.
\newblock A simple neural network module for relational reasoning.
\newblock In {\em Advances in neural information processing systems}, pages
  4967--4976, 2017.

\bibitem{shen2017deep}
F.~Shen, X.~Gao, L.~Liu, Y.~Yang, and H.~T. Shen.
\newblock Deep asymmetric pairwise hashing.
\newblock In {\em Proceedings of the 25th ACM international conference on
  Multimedia}, pages 1522--1530. ACM, 2017.

\bibitem{shen2015supervised}
F.~Shen, C.~Shen, W.~Liu, and H.~Tao~Shen.
\newblock Supervised discrete hashing.
\newblock In {\em Proceedings of the IEEE conference on computer vision and
  pattern recognition}, pages 37--45, 2015.

\bibitem{simonyan2014two}
K.~Simonyan and A.~Zisserman.
\newblock Two-stream convolutional networks for action recognition in videos.
\newblock In {\em Advances in neural information processing systems}, pages
  568--576, 2014.

\bibitem{simonyan2014very}
K.~Simonyan and A.~Zisserman.
\newblock Very deep convolutional networks for large-scale image recognition.
\newblock {\em arXiv preprint arXiv:1409.1556}, 2014.

\bibitem{song2019temporal}
X.~Song, C.~Lan, W.~Zeng, J.~Xing, X.~Sun, and J.~Yang.
\newblock Temporal-spatial mapping for action recognition.
\newblock {\em IEEE Transactions on Circuits and Systems for Video Technology},
  2019.

\bibitem{soomro2012ucf101}
K.~Soomro, A.~R. Zamir, and M.~Shah.
\newblock Ucf101: A dataset of 101 human actions classes from videos in the
  wild.
\newblock {\em arXiv preprint arXiv:1212.0402}, 2012.

\bibitem{sun2015human}
L.~Sun, K.~Jia, D.-Y. Yeung, and B.~E. Shi.
\newblock Human action recognition using factorized spatio-temporal
  convolutional networks.
\newblock In {\em Proceedings of the IEEE International Conference on Computer
  Vision}, pages 4597--4605, 2015.

\bibitem{sun2018optical}
S.~Sun, Z.~Kuang, L.~Sheng, W.~Ouyang, and W.~Zhang.
\newblock Optical flow guided feature: a fast and robust motion representation
  for video action recognition.
\newblock In {\em Proceedings of the IEEE conference on computer vision and
  pattern recognition}, pages 1390--1399, 2018.

\bibitem{tran2015learning}
D.~Tran, L.~Bourdev, R.~Fergus, L.~Torresani, and M.~Paluri.
\newblock Learning spatiotemporal features with 3d convolutional networks.
\newblock In {\em Proceedings of the IEEE international conference on computer
  vision}, pages 4489--4497, 2015.

\bibitem{tran2017convnet}
D.~Tran, J.~Ray, Z.~Shou, S.-F. Chang, and M.~Paluri.
\newblock Convnet architecture search for spatiotemporal feature learning.
\newblock {\em arXiv preprint arXiv:1708.05038}, 2017.

\bibitem{tran2018closer}
D.~Tran, H.~Wang, L.~Torresani, J.~Ray, Y.~LeCun, and M.~Paluri.
\newblock A closer look at spatiotemporal convolutions for action recognition.
\newblock In {\em Proceedings of the IEEE Conference on Computer Vision and
  Pattern Recognition}, pages 6450--6459, 2018.

\bibitem{varol2018long}
G.~Varol, I.~Laptev, and C.~Schmid.
\newblock Long-term temporal convolutions for action recognition.
\newblock {\em IEEE transactions on pattern analysis and machine intelligence},
  40(6):1510--1517, 2018.

\bibitem{wang2013action}
H.~Wang and C.~Schmid.
\newblock Action recognition with improved trajectories.
\newblock In {\em Proceedings of the IEEE international conference on computer
  vision}, pages 3551--3558, 2013.

\bibitem{wang2018learning}
J.~Wang and A.~Cherian.
\newblock Learning discriminative video representations using adversarial
  perturbations.
\newblock In {\em Proceedings of the European Conference on Computer Vision
  (ECCV)}, pages 685--701, 2018.

\bibitem{wang2018video}
J.~Wang, A.~Cherian, F.~Porikli, and S.~Gould.
\newblock Video representation learning using discriminative pooling.
\newblock In {\em Proceedings of the IEEE Conference on Computer Vision and
  Pattern Recognition}, pages 1149--1158, 2018.

\bibitem{wang2015action}
L.~Wang, Y.~Qiao, and X.~Tang.
\newblock Action recognition with trajectory-pooled deep-convolutional
  descriptors.
\newblock In {\em Proceedings of the IEEE conference on computer vision and
  pattern recognition}, pages 4305--4314, 2015.

\bibitem{wang2016temporal}
L.~Wang, Y.~Xiong, Z.~Wang, Y.~Qiao, D.~Lin, X.~Tang, and L.~Van~Gool.
\newblock Temporal segment networks: Towards good practices for deep action
  recognition.
\newblock In {\em European Conference on Computer Vision}, pages 20--36.
  Springer, 2016.

\bibitem{wang2018two}
X.~Wang, L.~Gao, P.~Wang, X.~Sun, and X.~Liu.
\newblock Two-stream 3-d convnet fusion for action recognition in videos with
  arbitrary size and length.
\newblock {\em IEEE Transactions on Multimedia}, 20(3):634--644, 2018.

\bibitem{wang2018non}
X.~Wang, R.~Girshick, A.~Gupta, and K.~He.
\newblock Non-local neural networks.
\newblock In {\em Proceedings of the IEEE Conference on Computer Vision and
  Pattern Recognition}, pages 7794--7803, 2018.

\bibitem{wu2015modeling}
Z.~Wu, X.~Wang, Y.-G. Jiang, H.~Ye, and X.~Xue.
\newblock Modeling spatial-temporal clues in a hybrid deep learning framework
  for video classification.
\newblock In {\em Proceedings of the 23rd ACM international conference on
  Multimedia}, pages 461--470. ACM, 2015.

\bibitem{xu2017learning}
D.~Xu, W.~Ouyang, E.~Ricci, X.~Wang, and N.~Sebe.
\newblock Learning cross-modal deep representations for robust pedestrian
  detection.
\newblock In {\em Proceedings of the IEEE Conference on Computer Vision and
  Pattern Recognition}, pages 5363--5371, 2017.

\bibitem{yao2015describing}
L.~Yao, A.~Torabi, K.~Cho, N.~Ballas, C.~Pal, H.~Larochelle, and A.~Courville.
\newblock Describing videos by exploiting temporal structure.
\newblock In {\em Proceedings of the IEEE international conference on computer
  vision}, pages 4507--4515, 2015.

\bibitem{yue2015beyond}
J.~Yue-Hei~Ng, M.~Hausknecht, S.~Vijayanarasimhan, O.~Vinyals, R.~Monga, and
  G.~Toderici.
\newblock Beyond short snippets: Deep networks for video classification.
\newblock In {\em Proceedings of the IEEE conference on computer vision and
  pattern recognition}, pages 4694--4702, 2015.

\bibitem{zhou2018temporal}
B.~Zhou, A.~Andonian, A.~Oliva, and A.~Torralba.
\newblock Temporal relational reasoning in videos.
\newblock In {\em Proceedings of the European Conference on Computer Vision
  (ECCV)}, pages 803--818, 2018.

\bibitem{zhou2016learning}
B.~Zhou, A.~Khosla, A.~Lapedriza, A.~Oliva, and A.~Torralba.
\newblock Learning deep features for discriminative localization.
\newblock In {\em Proceedings of the IEEE conference on computer vision and
  pattern recognition}, pages 2921--2929, 2016.

\bibitem{zhu2018end}
J.~Zhu, Z.~Zhu, and W.~Zou.
\newblock End-to-end video-level representation learning for action
  recognition.
\newblock In {\em 2018 24th International Conference on Pattern Recognition
  (ICPR)}, pages 645--650. IEEE, 2018.

\bibitem{zhu2016key}
W.~Zhu, J.~Hu, G.~Sun, X.~Cao, and Y.~Qiao.
\newblock A key volume mining deep framework for action recognition.
\newblock In {\em Proceedings of the IEEE Conference on Computer Vision and
  Pattern Recognition}, pages 1991--1999, 2016.

\end{thebibliography}

\end{document}